# BM3D VS 2-LAYER ONN

*Junaid Malik [a], Serkan Kiranyaz [b], Mehmet Yamac [c], Moncef Gabbouj [a]*

[a] Tampere Universities, Tampere, Finland
[b] Qatar University, Doha, Qatar
[c] HUAWEI Technologies, Tampere, Finland

**ABSTRACT**

Despite their recent success on image denoising, the need for deep and complex architectures still hinders the practical usage of CNNs. Older but computationally more efficient methods such as BM3D remain a popular choice, especially in resource-constrained scenarios. In this study, we aim to find out whether compact neural networks can learn to produce competitive results as compared to BM3D for AWGN image denoising. To this end, we configure networks with only two hidden layers and employ different neuron models and layer widths for comparing the performance with BM3D across different AWGN noise levels. Our results conclusively show that the recently proposed self-organized variant of operational neural networks based on a generative neuron model (Self-ONNs) is not only a better choice as compared to CNNs, but also provide competitive results as compared to BM3D and even significantly surpass it for high noise levels.

*Index Terms*— Image denoising, operational neural networks, self-organized operational neural networks, discriminative learning

## 1. INTRODUCTION

Image denoising is one of the most critical low-level imaging tasks because noise is inherent to all image acquisition processes. As obtaining a noise-free image is an ill-posed problem, prevalent denoising techniques resort to making certain assumptions about the nature of noise. The most common approach in this regard is to model the noise as an additive Gaussian process. Denoising images corrupted with additive white Gaussian noise (AWGN) has been a center of attention for decades. Early efforts considered denoising as an averaging process [1], either in frequency or spatial domain. These were succeeded by the non-local class of methods [2][3] which exploited the non-local similarity of patches within an image. BM3D [3] holds a pioneering status in such methods and is still competitive in challenging denoising tasks [4]. It remained the state-of-the-art method in AWGN denoising, until the advent of learning-based methods.

Discriminative learning-based denoising methods aim to learn a direct mapping from corrupted to clean images using annotated training samples. Artificial neural networks (ANNs) have become a standard choice for such problems and were shown in [5] to produce competitive denoising performance as compared to BM3D, given sufficient training examples and model complexity. Since then, most of the discriminative denoising methods make use of convolutional neural networks (CNNs), which are a special class of ANNs specialized for grid-structured data [6]. CNNs for image denoising is generally very deep [7], consisting of dozens of layers and a large number of trainable parameters generally over a million. The performance of CNN-based methods has been observed to be directly correlated with architectural complexity [8]. The enormous computational complexity is often neglected, especially since the performance of CNNs has been optimized quite well for specialized hardware devices such as GPUs and TPUs [9]. However, usage of such models is still expensive, and in many cases impractical, especially in the case of embedded devices that have limited computational resources.

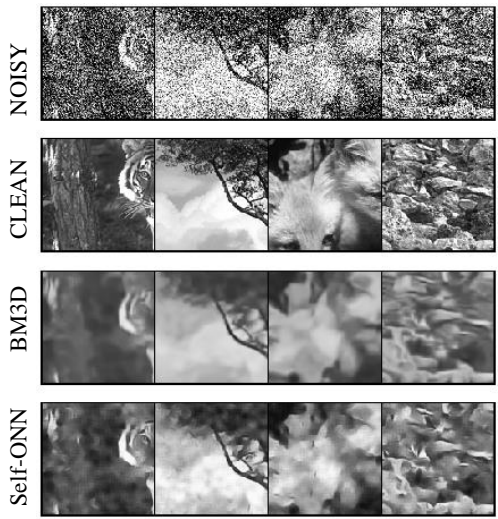

Figure 1. Visual results showing denoising performance of BM3D vs compact networks used in this study. The example noisy image is corrupted with $\sigma = 90$ AWGN noise.

While considerable efforts have been made to make CNNs more efficient [10], [11], recent studies have attributed the need for deep CNNs towards the inherent linear transformation based on MLPs.[12] A generalization of CNNs called the Operational Neural Networks (ONNs) [12] was proposed as a remedy which advocated the incorporation of patch-wise non-linearity, enabling increased heterogeneity among the neurons. However, unlike CNNs, ONNs requires careful curation of non-linear transformation candidates and additional training runs for optimizing the operator assignment. To address this drawback, a recent variant of ONNs called the self-organized operational neural network (Self-ONN) was proposed [13]. In a study on severe image restoration problems [14], Self-ONNs were observed to consistently outperform CNNs with compact network architectures and scarce training resources. This provided evidence that shallow neural networks can be employed for practical image denoising, given an appropriate neuron model such as the generative model of Self-ONN.

Amidst the ever-growing complexity of CNN-based denoising networks, in this study, we aim to shift the focus towards compact architectures and investigate their ability to discriminatively learn to denoise images. To this end, we configure networks with only two hidden layers, using both convolutional neurons (CNNs) and generative neurons (Self-ONNs), and compare their performance with BM3D for AWGN denoising. The experiments conducted span a variety of noise levels, layer widths, and degrees of non-linearity. Results show that compact networks powered by the generative neuron model outperform the convolutional networks and provide competitive results as compared to BM3D. The rest of the paper is structured as follows. Section 2 provides a brief introduction of related denoising works, Section 3 introduces the Self-ONNs and the underlying generative neuron model, Section 4 provides experimental details, Section 5 provides the results and discusses the key findings, and Section 6 concludes the paper.

## 2. RELATED WORK

The earliest work on image denoising treated the problem as an averaging problem either in the spatial or frequency domain. The most successful and widely used non-discriminative methods make use of the non-local self-similarity inherent to natural images. The non-local means (NLM) [2] method made use of this principle by replacing the traditional local means approach with a non-localized mean by weighing all other pixels based on their similarity with the target pixel. Later, the BM3D [3] method was proposed which also exploited the non-local similarity. It involved stacking similar non-local patches from within the image and applying collaborative filtering in a 3D transform domain. The method achieved great success and remained state-of-the-art for AWGN. With the advent of artificial neural networks (ANNs), a new class of methods emerged based on discriminative learning. Such methods learn to produce clean images (or residual images) directly from the noisy image. In one of the earliest such efforts [5], a 4-layer MLP consisting of approximately 8000 neurons and trained on more than 360 million training samples achieved comparable performance to that of BM3D. Later and most recent methods almost exclusively make use of CNNs. In [6], a 17-layer CNN was trained to around 140k patches to achieve state-of-the-art performance on different benchmarks. In [15], another deep CNN architecture was proposed that employed a noise variance map supplemented to the input image to improve denoising on spatially variant noise. Similarly, studies in [16] and [17] proposed different modifications to improve denoising such as batch renormalization, skip connections, and dilated convolutions. Generally, the focus has remained on deeper models with increasing model complexity to achieve better denoising performance. In the realm of compact architectures, the authors of [12] proposed ONNs, a new network model based on the so-called 'operational neurons', to remove the need for deep networks. A self-organized variant of ONNs, Self-ONN [13][14], has been proposed recently to remedy the practical problems with operator assignment in ONNs. In [14], Self-ONNs were shown to outperform traditional CNNs across various noise levels and types with compact networks and severe training constraints.

## 3. SELF-ORGANIZED OPERATIONAL NEURAL NETWORKS

### 3.1. Preliminaries

The 2D convolution operation between the $k^{th}$ neuron in the $l^{th}$ layer and $i^{th}$ neuron in the $(l-1)^{th}$ layer of a CNN can be formalized as follows:

$$x_{ik}^l(m,n) = \sum_{r=0}^{K-1}\sum_{t=0}^{K-1} w_{ik}^l(r,t) y_i^{l-1}(m+r, n+t) \quad (1)$$

Here, $w_{ik} \in \mathbb{R}^{K \times K}$ is the kernel connecting the two neurons, while $x_{ik}^l \in \mathbb{R}^{M \times N}$ is the input map, and $y_i^{l-1} \in \mathbb{R}^{M \times N}$ are the $l^{th}$ and $(l-1)^{th}$ layers' $k^{th}$ and $i^{th}$ neurons' outputs respectively. An operational neuron generalizes the above convolution operation as follows:

$$\overline{x_{ik}^l}(m,n) = P_k^l \left( \psi_k^l \left( w_{ik}^l(r,t), y_i^{l-1}(m+r,n+t) \right) \right)_{(r,t)=(0,0)}^{(K-1,K-1)} \quad (2)$$

where $\psi_l^k(\cdot): \mathbb{R}^{MN \times K^2} \to \mathbb{R}^{MN \times K^2}$ and $P_k^l(\cdot): \mathbb{R}^{MN \times K^2} \to \mathbb{R}^{MN}$ are the so-called *nodal* and *pool* functions, respectively, assigned to the $k^{th}$ neuron of $l^{th}$ layer. A practical issue when deploying ONNs is the selection of candidate operators and optimization of operator assignment, which is needed before the ONN can be trained. This introduces subjectivity as the pre-selection of operators can bias the eventual performance. The motivation behind Self-ONNs is to use a generative neuron model that can remedy the fore-mentioned issues with ONN.

## 3.2. Generative Neurons

In Self-ONNs, the need for manual curation of operators is alleviated by introducing the idea of generative neurons, which utilize Taylor-series based function approximation to *generate* optimal non-linear functions. Specifically, in terms of notation used in (2), the nodal transformation of a generative neuron takes the following general form:

$$\widetilde{\psi_k^l}\left(w_{ik}^{l(Q)}(r,t), y_i^{l-1}(m+r, n+t)\right)$$
$$= \sum_{q=1}^{Q} w_{ik}^{l(Q)}(r, t, q) \left(y_i^{l-1}(m+r, n+t)\right)^q \quad (3)$$

In (3), $Q$ is a hyperparameter which controls the degree of the Taylor polynomial approximation, and in turn the extent of non-linearity, and $w_{ik}^{l(Q)}$ is a learnable parameter. A key difference in (3) as compared to the convolutional (1) and operational (2) model is the need for $Q$ times additional parameters. Therefore, $w_{ik} \in \mathbb{R}^{K \times K}$ is replaced by $w_{ik}^{l(Q)} \in \mathbb{R}^{K \times K \times Q}$. The output of the generative neuron, $\tilde{x}_{ik}^l$ can now be expressed as,

$$\widetilde{x_{ik}^l}(m,n) =$$
$$P_k^l \left( \sum_{q=1}^{Q} w_{ik}^{l(Q)}(r,t,q) \left(y_i^{l-1}(m+r, n+t)\right)^q \right)_{(r,t)=(0,0)}^{(K-1,K-1)} \quad (4)$$

During the training of the network, $w_{ik}^{l(Q)}$ is constantly updated based on the problem at hand, and results in novel non-linear transformations. This not only alleviates the need for defining operator candidates but also enables heterogeneous operator assignment as each neuron can effectively optimize its individual connections on-the-fly and within the same training run, thus promoting inter-neuronal diversity.

## 3.3. Representation in Terms of Convolution

A specific case of formulation (4) can also be expressed in terms of the widely applicable convolutional model. If the pooling operator $P_k^l$ is fixed to summation operator, $\tilde{x}_{ik}^l$ can then defined as:

$$\widetilde{x_{ik}^l}(m)$$
$$= \sum_{r=0}^{K-1} \sum_{t=0}^{K-1} \sum_{q=1}^{Q} w_{ik}^{l(Q)}(r,t,q) \left(y_i^{l-1}(m+r, n+t)\right)^q \quad (5)$$

Using (1), the formula in (5) can be further simplified as follows:

$$\widetilde{x_{ik}^l} = \sum_{q=1}^{Q} Conv2D\left(w_{ik}^{l(Q)}, \left(y_i^{l-1}\right)^q\right) \quad (6)$$

Hence, the formulation can be accomplished by applying $Q$ 2D convolution operations. If $Q$ is set to 1, (6) entails the convolutional formulation of (1). Therefore, as CNN is a subset of ONN corresponding to a specific operator set, it is also a special case of Self-ONN with $Q = 1$ for all neurons.

## 4. EXPERIMENTS

### 4.1. Network Architecture

We employ compact networks with only two-hidden layers for the color image denoising problem. The number of neurons in each layer (width) is varied within the range of [64,128]. For each architecture, both convolutional neurons and generative neurons are used and evaluated. For Self-ONNs (networks using generative neurons), different values of hyperparameter $Q$ are tested, within the range [3,5,7]. The kernel sizes are fixed to $3 \times 3$ for all networks. The naming convention used is CNN-$X$ and Self-ONN-$Q$-$X$ where $X$ is the number of neurons in the hidden layers.

### 4.2. Datasets

For training, we use 200k patches of size 40x40 extracted from the 400 images of the BSD400 dataset, similar to [6]. For testing, we use 68 images from BSD68 [18] dataset, 24 images from Kodak [19], and 18 images from McMaster [20] dataset. All test images are high-quality grayscale images with a minimum resolution of 480x320 pixels. For noisy data generation, random AWGN noise with sigma levels in the range [30,60,90] is generated and added to the clean images to generate their noisy counterparts. Noisy images are further clipped to a range of [0,1] to achieve a standardized comparison.

### 4.3. Training Settings

The training to validation ratio of the training data is set to 95:5. All the networks are trained for minimizing L2-loss using Adam [21] optimization with an initial learning rate of $1e-3$ for 100 epochs, after which the model achieving the highest validation performance is chosen. For evaluation, we use the peak signal-to-noise ratio (PSNR) measure. All experiments were conducted in Python using the FastONN library [22].

## 5. RESULTS AND DISCUSSION

### 5.1. Comparison with BM3D

The Self-ONNs achieve a better performance than CNNs for all network sizes and noise levels and enable the compact networks to be more competitive with BM3D, as shown in Figure 2 and Table 1. For $\sigma = 30$, the best Self-ONN architecture, Self-ONN-7-128 achieves an improvement of 0.19dB. For higher noise levels, the gap between Self-ONNs and BM3D widens even more. On average, Self-ONN-3-128 surpasses BM3D by 0.6dB when $\sigma = 60$ and Self-ONN-7-128 outperforms BM3D by 1.3dB when $\sigma = 90$, across the three datasets. Overall, across the three datasets, the average best performance of 25.57dB is achieved by the Self-ONN-7-128 network, surpassing BM3D by 0.7dB. The higher performance of Self-ONN reaffirms the rationale that the limited non-linearity posed by the CNNs is not sufficient to achieve a decent denoising performance.

Table 1. Comparison of CNN and Self-ONN networks with BM3D across different noise levels.

|  | KODAK | | | McMaster | | | CBSD68 | | |
|---|---|---|---|---|---|---|---|---|---|
|  | σ=30 | σ=60 | σ=90 | σ=30 | σ=60 | σ=90 | σ=30 | σ=60 | σ=90 |
| BM3D | **28.58** | 25.05 | 22.44 | 29.30 | 24.76 | 21.59 | 27.17 | 23.65 | 21.27 |
| CNN-64 | 28.43 | 25.01 | 23.08 | 29.29 | 25.69 | 23.55 | 27.45 | 24.26 | 22.40 |
| CNN-128 | 28.47 | 25.08 | 23.11 | 29.28 | 25.73 | 23.55 | 27.47 | 24.28 | 22.43 |
| Self-ONN-3-64 | 28.54 | 25.09 | 23.12 | 29.39 | 25.77 | **23.66** | 27.55 | 24.31 | 22.47 |
| Self-ONN-3-128 | 28.55 | 25.12 | 23.10 | 29.41 | **25.82** | 23.59 | 27.56 | 24.33 | 22.45 |
| Self-ONN-5-64 | 28.56 | 25.10 | 23.12 | 29.40 | 25.81 | 23.65 | 27.55 | **24.34** | 22.46 |
| Self-ONN-5-128 | 28.58 | 25.12 | 23.11 | 29.39 | 25.76 | 23.64 | 27.56 | 24.32 | 22.47 |
| Self-ONN-7-64 | 28.54 | 25.12 | **23.12** | 29.40 | 25.77 | 23.64 | 27.56 | 24.33 | 22.46 |
| Self-ONN-7-128 | 28.57 | **25.13** | 23.12 | **29.47** | 25.76 | 23.65 | **27.57** | 24.33 | **22.48** |

Using a more potent generative neuron model results in better performance levels and enables the compact networks to surpass BM3D by a considerable margin.

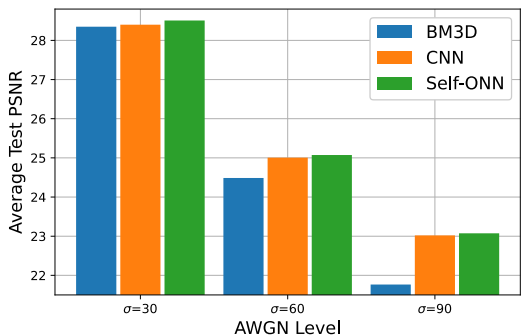

Figure 2. The average performance of BM3D, CNNs, and Self-ONNs for different noise levels.

### 5.2. Effect of nonlinearity on denoising performance

The overall top network, SelfONN-7-128, has $Q$ set to the highest value of 7. Table 2 shows the performance improvement when increasing the value of $Q$. We see that irrespective of the width, going from $Q = 1$ (CNN) to Self-ONN ($Q = 3$) results in consistent improvement of as much as 0.35% in the case of the network with 64 neurons. This again provides evidence in favor of the enhanced non-linearity achieved by the generative neuron model of Self-ONNs. Increasing the value of $Q$ beyond 3 generally results in a slight decrease in performance for the network with 64 neurons. This can be attributed to the fact that the size of the training set can be a potential bottleneck and increasing the non-linearity beyond a certain flexing point can result in overfitting, especially when the network complexity is low. When the number of intermediate neurons is increased from 64 to 128, the effect of increasing $Q$ is more pronounced. Interestingly, for $\sigma = 90$, the rise in generalization performance when increasing $Q$ from 3 to 5 is even more than the one observed when $Q$ is increased from 1 to 3. A reason for this can be the ability of generative neurons to spawn novel non-linear functions based on the learning problem. As the generation of function profiles is not constrained, it is possible that increasing the value of $Q$ from 3 to 5 enabled the generative neuron to create a unique and superior non-linear function which was not possible with $Q < 5$. This re-emphasizes the need for more granular heterogenous adjustment of $Q$ values to achieve an optimal trade-off between the enhanced non-linearity and the increase in trainable parameters.

Table 2. Impact of increasing non-linearity on the denoising performance of compact networks.

| Neurons | ΔQ | Percentage change in PSNR | | |
|---|---|---|---|---|
|  |  | σ=30 | σ=60 | σ=90 |
| 64 | 1→3 | 0.35 | 0.28 | 0.31 |
|  | 3→5 | 0.04 | 0.10 | -0.04 |
|  | 5→7 | -0.01 | -0.05 | -0.01 |
| 128 | 1→3 | 0.34 | 0.25 | 0.07 |
|  | 3→5 | 0.00 | -0.09 | 0.13 |
|  | 5→7 | 0.12 | 0.03 | 0.04 |

### 6. CONCLUSION

The efficacy of compact neural network architectures for solving the image denoising problem was investigated by comparing them with the state-of-the-art method of BM3D, using a synthetic AWGN noise model with various degrees of corruption. Moreover, both convolutional and generative neuron models were evaluated. Models with a larger number of neurons performed better as compared to relatively slimmer networks. The networks utilizing generative neurons, Self-ONNs, outperform CNN across all datasets and noise levels. Self-ONNs also provide competitive performance as compared to BM3D and outperform it across all noise levels. Increasing the amount of non-linearity in Self-ONNs generally results in better performance but makes the network more prone to overfitting and can potentially lead to worse results, especially with smaller networks. Optimizing the level of non-linearity of Self-ONNs may lead to better performance and should be investigated further. This will also be a focus of our future work.